\documentclass[a4paper,fleqn]{cas-dc}
\usepackage[numbers]{natbib}
\usepackage{graphicx}
\usepackage{amsmath}
\usepackage{amssymb}
\usepackage{booktabs}
\usepackage{multirow}
\usepackage{xcolor}
\usepackage{subcaption}
\usepackage{hyperref}
\hypersetup{
    colorlinks=true,
    linkcolor=red,
    filecolor=magenta,      
    urlcolor=magenta,
    citecolor=[RGB]{54 125 190}, 
}

\def\tsc#1{\csdef{#1}{\textsc{\lowercase{#1}}\xspace}}
\tsc{WGM}
\tsc{QE}

\begin{document}
\let\WriteBookmarks\relax
\def\floatpagepagefraction{1}
\def\textpagefraction{.001}

\shorttitle{Weakly Supervised Tooth-marked Tongue Recognition}    

\shortauthors{Zhang et al.}  

\title [mode = title]{Weakly Supervised Object Detection for Automatic Tooth-marked Tongue Recognition}  

\author[1]{Yongcun Zhang}
\affiliation[1]{organization={Department of Artificial Intelligence, Xiamen University},
            city={Xiamen},
            postcode={361005}, 
            state={Fujian},
            country={China}} 
\author[2]{Jiajun Xu}
\author[1]{Yina He}
\author[1]{Shaozi Li}
\author[1]{Zhiming Luo}
\author[3]{Huangwei Lei}\cormark[1]

\affiliation[2]{organization={Pediatrics of TCM of Women and Children’s Hospital, School of Medicine, Xiamen University},
 city={Xiamen},
                postcode={361005}, 
                state={Fujian},
                country={China}}

\affiliation[3]{organization={College of Traditional Chinese Medicine, Fujian University of Traditional Chinese Medicine},
                city={Fuzhou},
                postcode={350122}, 
                state={Fujian},
                country={China}}

\cortext[1]{Corresponding author}

\begin{abstract}
Tongue diagnosis in Traditional Chinese Medicine (TCM) is a crucial diagnostic method that can reflect an individual's health status. Traditional methods for identifying tooth-marked tongues are subjective and inconsistent because they rely on practitioner experience. We propose a novel fully automated \textbf{W}eakly \textbf{S}upervised method using \textbf{V}ision transformer and \textbf{M}ultiple instance learning (\textbf{WSVM}) for tongue extraction and tooth-marked tongue recognition. Our approach first accurately detects and extracts the tongue region from clinical images, removing any irrelevant background information. Then, we implement an end-to-end weakly supervised object detection method. We utilize Vision Transformer (ViT) to process tongue images in patches and employ multiple instance loss to identify tooth-marked regions with only image-level annotations. WSVM achieves high accuracy in tooth-marked tongue classification, and visualization experiments demonstrate its effectiveness in pinpointing these regions. This automated approach enhances the objectivity and accuracy of tooth-marked tongue diagnosis. It provides significant clinical value by assisting TCM practitioners in making precise diagnoses and treatment recommendations. Code is available at \url{https://github.com/yc-zh/WSVM}.
\end{abstract}

\begin{keywords}
Tooth-marked tongue \sep Weakly supervised \sep Tongue recognition
\end{keywords}

\maketitle

\section{Introduction}
\label{sec:intro}

\begin{figure}[t]
\vspace{1.5em}
    \centering
    \begin{subfigure}[t]{0.24\linewidth}
        \centering
        \includegraphics[width=\linewidth]{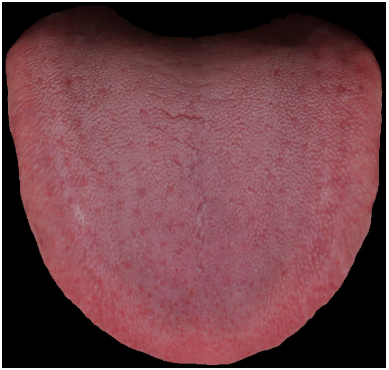}
        \caption{}
        \label{fig:a}
    \end{subfigure}
    \hfill
    \begin{subfigure}[t]{0.24\linewidth}
        \centering
        \includegraphics[width=\linewidth]{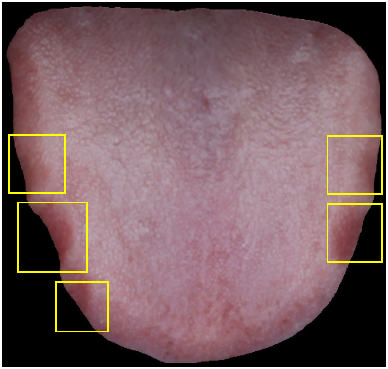}
        \caption{}
        \label{fig:b}
    \end{subfigure}
    \begin{subfigure}[t]{0.24\linewidth}
        \centering
        \includegraphics[width=\linewidth]{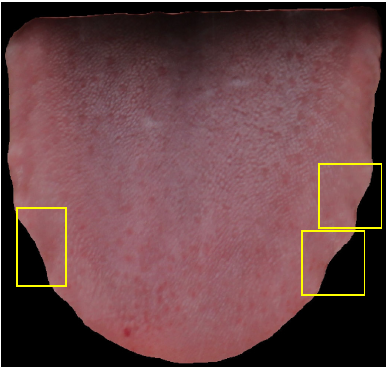}
        \caption{}
        \label{fig:c}
    \end{subfigure}
    \hfill
    \begin{subfigure}[t]{0.24\linewidth}
        \centering
        \includegraphics[width=\linewidth]{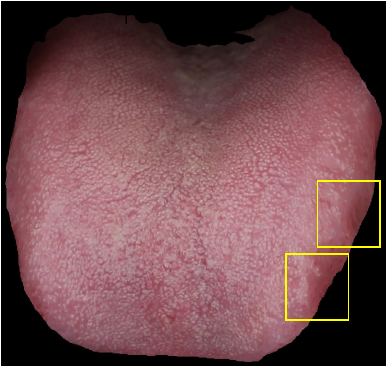}
        \caption{}
        \label{fig:d}
    \end{subfigure}
    \begin{subfigure}[t]{0.24\linewidth}
        \centering
        \includegraphics[width=\linewidth]{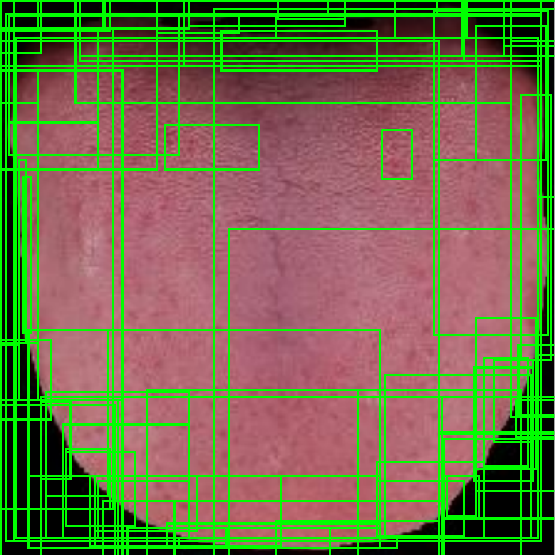}
        \caption{}
        \label{fig:e}
    \end{subfigure}
    \hfill
    \begin{subfigure}[t]{0.24\linewidth}
        \centering
        \includegraphics[width=\linewidth]{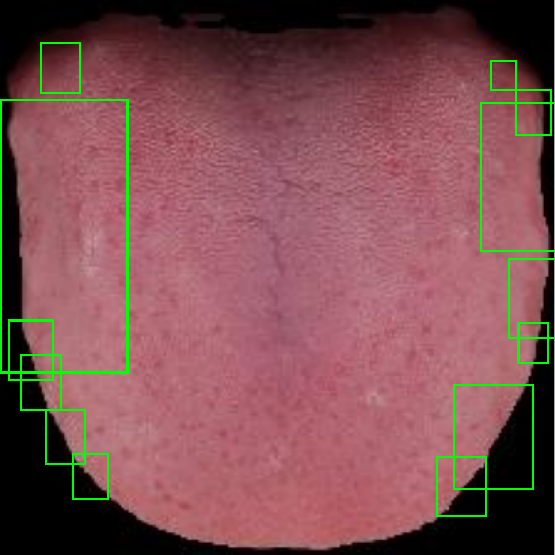}
        \caption{}
        \label{fig:f}
    \end{subfigure}
    \begin{subfigure}[t]{0.24\linewidth}
        \centering
        \includegraphics[width=\linewidth]{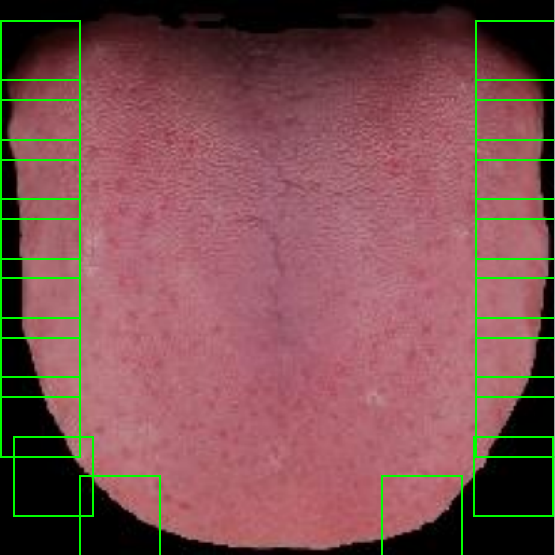}
        \caption{}
        \label{fig:g}
    \end{subfigure}
    \hfill
    \begin{subfigure}[t]{0.24\linewidth}
        \centering
        \includegraphics[width=\linewidth]{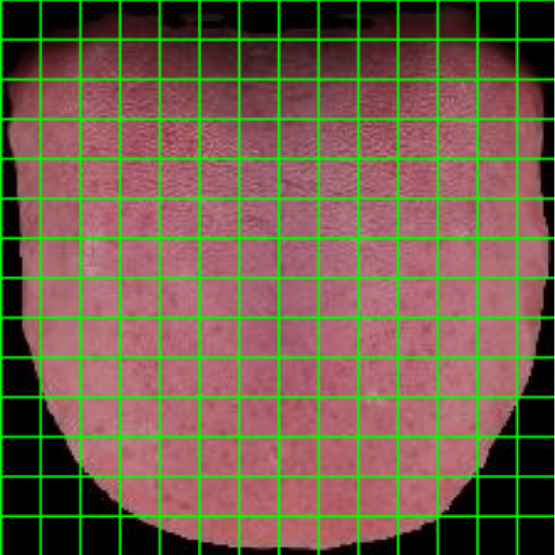}
        \caption{}
        \label{fig:h}
    \end{subfigure}
    \caption{~\ref{fig:a} is a normal tongue,~\ref{fig:b}~\ref{fig:c}~\ref{fig:d} are three different tooth-marked tongues,~\ref{fig:e}~\ref{fig:f}~\ref{fig:g}~\ref{fig:h} are regional proposal approaches for four different methods of identifying tooth marks and tongues }
    \label{fig:tongues}
   
\end{figure}

Tongue diagnosis is an important diagnostic method in Traditional Chinese Medicine (TCM) which can be used to identify an individual's health status~\cite{kirschbaum2000atlas,bing2010diagnostics,liu2023survey}. 
By examining the shape, color, and moisture of the tongue body and coating, practitioners can infer a person's health state~\cite{hassoon2024tongue,brown2024medical}. 
One representative symptom in this context is the ``tooth-marked tongue", as shown in Fig.~\ref{fig:tongues}. 
The yellow boxes indicate tooth marks, identified at the tongue's edges caused by tooth pressure~\cite{li2018tooth, shi2023ammonia}. 
In Fig.~\ref{fig:a}, the edge of a normal tongue is smooth and exhibits a full red color. 
In contrast, in Fig.~\ref{fig:b} a tongue with tooth marks often shows indentations from tooth pressure, which may be concave in shape (Fig.~\ref{fig:c}) or darker in color, creating a stark contrast with the surrounding red tissue(Fig.~\ref{fig:d}). 
TCM practitioners use these observations to make preliminary judgments about a patient's constitution, condition, and disease type, then formulate targeted treatment plans. 
Therefore, accurate recognition and analysis of the tooth-marked tongue hold significant clinical value~\cite{tang2020automatic, shao2014recognition, zhou2022weakly}.

The traditional diagnosis of a tooth-marked tongue heavily relies on the observational skills and experience of TCM practitioners, making the diagnosis highly subjective and potentially inconsistent~\cite{zhou2022weakly,li2018tooth,zhang2021computational,li2008towards,tania2019advances}.
With the advent of computer vision and artificial intelligence technologies~\cite{krizhevsky2012imagenet, he2016deep, ren2015faster, redmon2016you, he2024towards}, more research is being conducted to explore machine learning and deep learning methods for the automated recognition and analysis of the tooth-marked tongue~\cite{shao2014recognition, li2008towards, tania2019advances, li2018tooth, sun2019tooth, zhou2022weakly, he2024tsesnet}. 
These automated methods not only enhance the objectivity and accuracy of tooth-marked tongue diagnosis but also provide additional information to assist TCM practitioners in making more precise diagnoses and treatment recommendations.

However, these methods typically require a large amount of precisely annotated data for training, especially for labeling the regions of tooth marks. 
Weakly supervised learning has emerged as a feasible solution to this issue~\cite{bilen2016weakly, tang2017multiple, zhou2018brief, cheng2020high, wang2023alwod}. 
Weakly supervised methods use partially or coarsely labeled data (e.g., image-level labels) to train models, reducing the dependency on extensively annotated data.
This approach has already been successfully applied to other medical image analysis tasks, such as lesion detection and organ segmentation~\cite{afshari2019weakly, yang2020weakly, sun2023weakly}. 
Despite the existence of several effective weakly supervised methods for tooth-marked tongue recognition, these methods face several challenges.

\begin{figure}[t]
    \vspace{0.5em}
    \centering
    \includegraphics[width=1\linewidth]{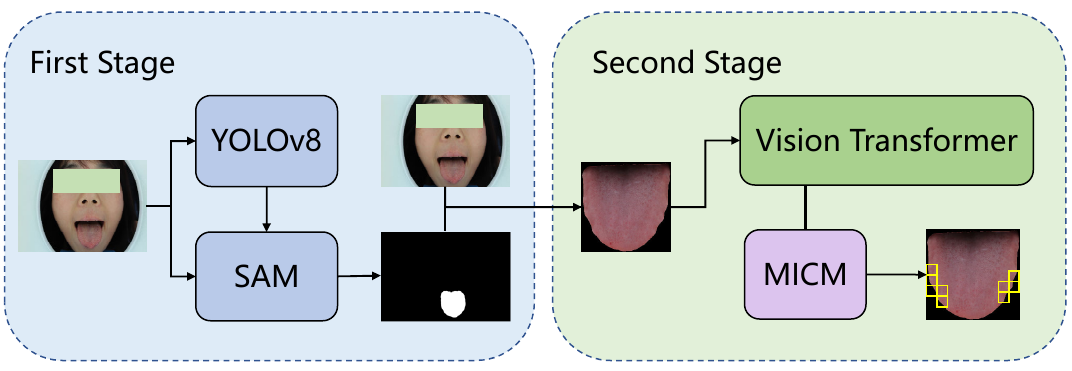}
    \caption{Overall Framework of Our Approach. Our approach includes the first stage of automatic tongue foreground extraction and the second stage of weakly supervised tooth-mark tongue detection.}
    \label{fig:overall}
\end{figure}

During the data processing stage, human intervention is required to manually annotate the tongue foreground in clinical facial images~\cite{zhou2022weakly, li2018tooth, tang2020automatic}. 
Recognizing tooth-marked tongues often involves multiple training stages, such as an independent region proposal stage~\cite{zhou2022weakly, li2018tooth} or the need to train multiple models. 
Typical examples of region proposal include the selective search shown in Fig.~\ref{fig:e}, the region generation based on tongue edge concavity proposed by Li et al.~\cite{li2018tooth}. shown in Fig.~\ref{fig:f}, and the spatial region generation proposed by Zhou et al.~\cite{zhou2022weakly} shown in Fig.~\ref{fig:g}. Among these methods, Selective search results in a significant computational burden, and Fig.~\ref{fig:f} is not effective in handling scalloped tongue with indistinct marginal depressions such as Fig.~\ref{fig:d}. 
On the other hand, an example of training multiple models is to first classify to enhance features and then further fine-tune for recognition~\cite{zhou2022weakly}. 
Additionally, these methods are designed for tooth-marked tongue recognition and can't be directly applied to other tasks, such as crack tongue recognition.

To overcome these issues, we propose a fully automated \textbf{W}eakly \textbf{S}upervised object detection method based on \textbf{V}ision transformer (ViT)~\cite{dosovitskiy2020image} and \textbf{M}ultiple instance learning (MIL) named WSVM as shown in Fig.~\ref{fig:overall}.
WSVM consists of two parts: (1) Tongue foreground extraction and (2) Weakly Supervised tooth-mark tongue detection. In the first stage, we use a YOLOv8~\cite{yolov8_ultralytics} and SAM~\cite{kirillov2023segment} to accurately identify the tongue region from complete facial images captured in clinical settings and extract an independent and complete foreground image of the tongue. By doing so, we can remove a large amount of irrelevant background information, allowing our model to focus on more valuable tongue features during the recognition phase.  In the second stage, we propose an end-to-end weakly supervised object detection method to recognize tooth-marked tongues. Our model is based on the Vision Transformer (ViT)~\cite{dosovitskiy2020image}, in which we treat the tongue image patches as region proposals and feed them into the transformer encoder, as shown in Fig.~\ref{fig:h}. Then, we remove the classification head in ViT, and consider the output features of each patch as an instance. Next, we pass them into a multiple instance calculation module (MICM) and employ the multiple instance loss function for training. Therefore, we can achieve weakly supervised object detection of the tooth-marked tongue by using only image-level annotations. Finally, we conduct experiments on a primary dataset and a public dataset to evaluate the performance of our proposed methods.

Our main contributions are as follows:
\begin{itemize}
\item We develop a fully automated method for tongue detection, capable of accurately extracting the tongue foreground from raw clinical images.
\item We propose an end-to-end weakly supervised object detection method based on Vit and MIL, which can recognize tooth-marked tongues using only image-level labels.
\item Through experiments, we validated the effectiveness of WSVM in detecting tongues and distinguishing and recognizing tooth-marked tongues.
\end{itemize}

\begin{figure*}
    \centering
    \includegraphics[width=0.96\linewidth]{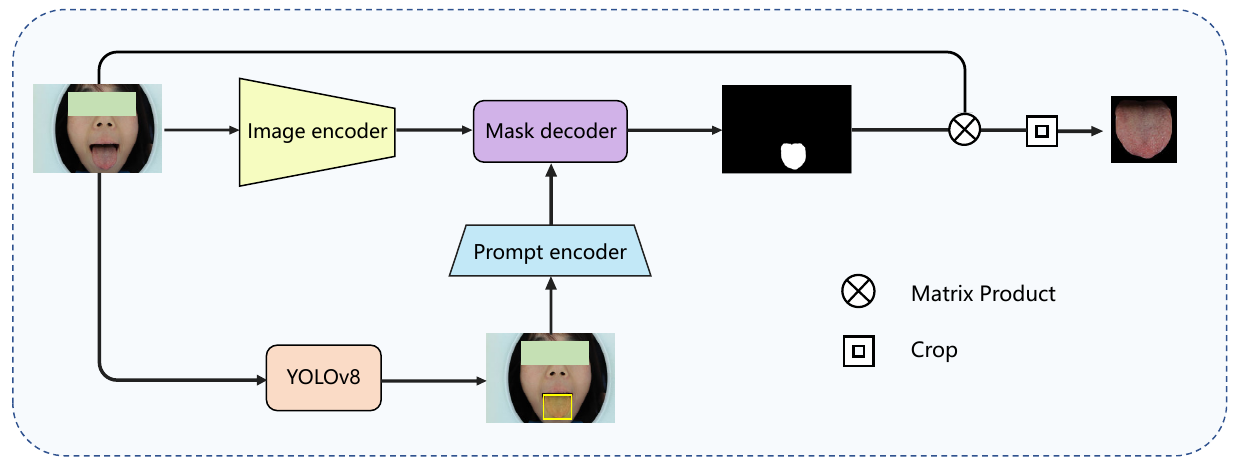}
    \caption{Fully automatic tongue extraction. It used SAM's zero-shot segmentation capability. 
    We start with bounding boxes from our trained YOLOv8n~\cite{yolov8_ultralytics} tongue detector as prompts for SAM to generate a tongue mask. 
    This mask is then multiplied with the original image to remove the background, isolating the tongue. The segmented tongue region is then cropped for further analysis.}
    \label{fig:detect}
\end{figure*}

\section{Related works}
The advancement of automatic tongue diagnosis technology has improved the objective assessment of patients' conditions based on tongue characteristics~\cite{ hsu2010automatic, tania2019advances, hu2019automated, vocaturo2020machine, QIAO2024106643, QIU2023104271}, particularly focusing on the study of tooth-marked tongues. In the study of tooth-marked tongues, some researchers have analyzed tongue characteristics related to tooth marks to aid in their identification.
Building on this, several works have employed computer vision methods to classify whether a tongue has tooth marks. 
However, these methods do not fully meet the clinical diagnostic requirements.
Therefore, works such as those by~\cite{weng2021weakly} and~\cite{shi2023ammonia} have developed methods to detect tooth-marked regions on the tongue. 
Annotating tooth-marked regions is time-consuming, labor-intensive, and subjective. 
Consequently, weakly supervised object detection methods, which require only image-level labels, have gained popularity among researchers. 
For example,~\cite{sun2019tooth} proposed a visual explanation method that uses convolutional neural networks (CNNs) to process the entire tongue image to extract features and employs gradient-weighted class activation mapping to highlight the tooth-marked areas. 
Wang et al.~\cite{wang2020artificial} used deeper CNNs to enhance feature extraction capabilities and achieve better recognition results. 
~\cite{li2018tooth} proposed a three-stage method to address the challenges of tooth-marked tongue diagnosis. 
This method first identifies suspected regions using concavity information, then extracts deep features with a CNN, and finally applies a multiple instance classifier for the final judgment. 
Nevertheless, multi-stage networks present challenges in training and clinical diagnosis interpretation and involve considerable redundant computation due to region proposals.~\cite{zhou2022weakly} introduced an end-to-end weakly supervised method for detecting tooth-marked tongue regions. However, manual segmentation of tongue regions, region proposals, and multiple training iterations are still required to improve recognition capabilities.

\section{Method}
Our WSVM consists of two stages, as shown in Fig.~\ref{fig:overall}. 
In the first stage, the tongue foreground is extracted from clinical full-face images to reduce background interference in recognizing tooth marks.
In the second stage, we use our proposed weakly supervised object detection method to classify the tongue and identify the tooth-marked regions.

\subsection{Preliminaries: Multiple instance learning}
Multiple instance learning (MIL) is a machine learning paradigm where labels are assigned to sets of instances (bags) rather than individual instances~\cite{maron1997framework, carbonneau2018multiple, herrera2016multiple, li2018tooth}. 
This approach is instrumental in medical imaging, where precise instance-level annotations are often unavailable or impractical to obtain. 
In an MIL framework, a positive bag contains at least one positive instance, while a negative bag contains only negative instances. 
MIL has been successfully applied to various medical imaging tasks, including tumor detection, histopathological image analysis, and retinal disease detection. 
MIL's flexibility allows models to learn from ambiguous and weakly labeled data, making it well-suited for tooth-marked tongue recognition, where exact tooth-marked locations may not be annotated.

In multiple instance learning, consider a training set \( \{(B_i, y_i)\}_{i=1}^N \), where \( B_i = \{x_{i1}, x_{i2}, \ldots, x_{im}\} \) represents a bag of instances, and \( y_i \in \{0, 1\} \) is the label of the bag. 
If \( y_i = 1 \), at least one positive instance exists in the bag \( B_i \); if \( y_i = 0 \), it indicates that all instances in the bag \( B_i \) are negative. 
A common assumption in MIL is the max-instance assumption, where the label of the bag is determined by its most representative instance. Based on this assumption, the probability of a bag being positive is defined as:
\begin{equation}
     P(y_i = 1 \mid B_i) = 1 - \prod_{j=1}^m (1 - p(x_{ij})) 
\end{equation}
Here, \( p(x_{ij}) \) represents the probability that instance \( x_{ij} \) is positive.

To train an MIL model, we use the following loss function:
\begin{equation}
\begin{aligned}
\mathcal{L} = 
& - \sum_{i=1}^N \left[ y_i \log P(y_i = 1 \mid B_i) \right. \\
& \left. + (1 - y_i) \log (1 - P(y_i = 1 \mid B_i)) \right]
\end{aligned}
\end{equation}
By minimizing this loss function, the model learns to extract effective features from ambiguous and weakly labeled data, thus performing well in MIL tasks.
In this paper, unlike the aforementioned approach that utilizes information from all instances, we leverage only the most positive instance for bag classification.
This strategy is particularly beneficial in scenarios where irrelevant or misleading data might otherwise obscure the true signals, ensuring that the model remains focused on the most informative aspects of the data.

\subsection{Fully automatic tongue extraction}

To eliminate the interference of other facial organs in recognizing tooth-marked tongues, we implemented a fully automatic tongue extraction method that removes the background from clinically taken images and extracts the tongue foreground. Our approach includes two main components: tongue detection and tongue segmentation. As illustrated in Fig.~\ref{fig:detect}
Our method ensures accurate and efficient tongue extraction, facilitating the subsequent tooth-marked tongue recognition.

\ \textbf{Tongue detection.} We use the YOLOv8n (You Only Look Once version 8 nano)~\cite{yolov8_ultralytics} model to get a high-quality box prompt for tongue detection. 
We train and test the YOLOv8n model on 224 annotated facial images to accurately detect the tongue's region of interest (ROI).
This allows the model to provide a bounding box around the tongue, facilitating efficient separation from other facial features and providing an accurate and reliable prompt for subsequent segmentation.

\textbf{Tongue mask extraction.} Once the tongue ROI is identified, we use the Segment Anything Model (SAM)~\cite{kirillov2023segment} for detailed segmentation within the bounding box.
SAM precisely delineates the tongue's boundaries, capturing even subtle features.
After segmentation, we multiply the original image by the mask to remove the background.
The tongue is then cropped from the image using the tongue mask, resulting in a clear foreground image without irrelevant background information.
This step is crucial as it allows subsequent analysis to focus solely on the tongue, thereby enhancing the accuracy of tooth-marked tongue recognition.

\begin{figure*}[t]
\begin{center}
\includegraphics[width=0.92\linewidth]{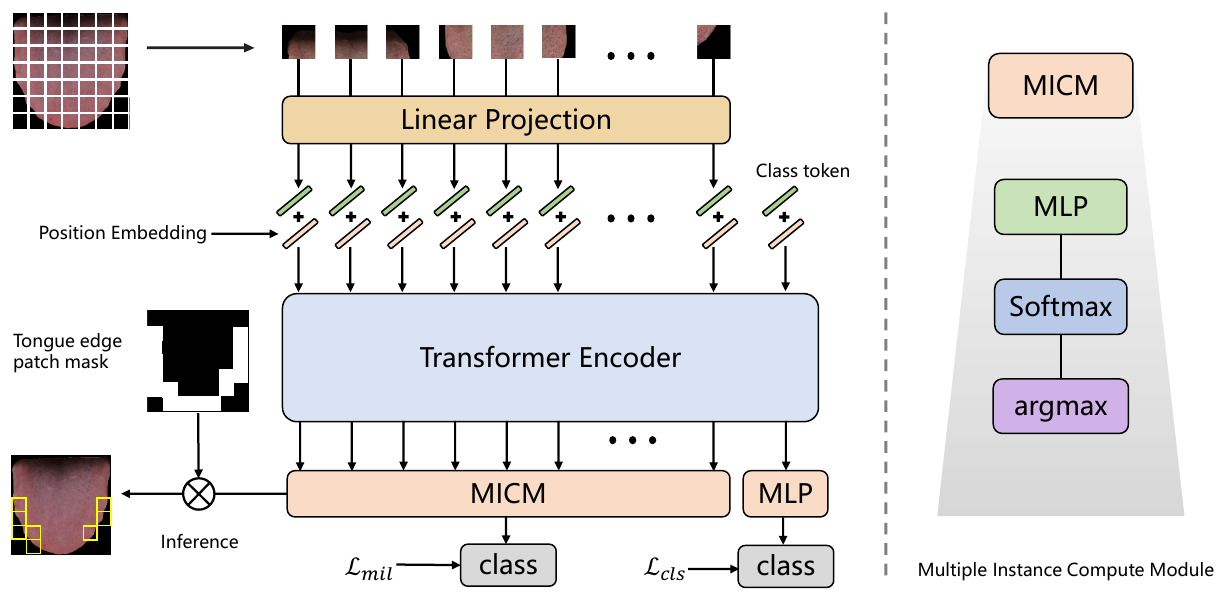}
\end{center}
\caption{Weakly Supervised Tooth-Marked Recognition.
We build our model depending on the ViT, incorporating a multiple instance calculation module and using weakly supervised loss along with image-level labels for supervision. }
\label{fig:recognition}
\end{figure*}

\subsection{Tooth-marked Tongue Recognition}
To recognize tooth-marked tongues, we design an end-to-end weakly supervised tooth-marked tongue recognition model based on ViT as shown in Fig.~\ref{fig:recognition}.
The input batch B images are first resized and normalized to a fixed resolution suitable for our model: $X \in R^{B \times C\times H \times W}$. 
The normalized image is then divided into smaller, non-overlapping patches. 
Each patch is $p\times p$ pixels, where $p$ is a 16 in this paper for the ViT base version we used. 
Thus, we get $N = \frac{H}{p} \times \frac{W}{p}$ patches from each image. 
These patches serve as region proposals for tooth-marked tongue recognition. 
Subsequently, the patches are projected to a higher-dimensional tensor $f_{in} \in  R ^{ B \times N \times D} $ using linear projection, where $D$ is the embedding dimension.
Positional encodings are added to the projected patch embeddings to incorporate spatial information. The $N$ embeddings and class token are then fed into the transformer encoder which captures complex dependencies and patterns within the image and outputs a feature embedding $f_{out} \in  R ^{ B \times (N+1) \times D} $. 
Instead of using the standard classification head of ViT, we propose a Multiple Instance Calculation Module (MICM). 
The transformer encoder outputs $B \times N$ feature vectors $x_i \in  R ^{D} $ and a class vector $x_cls \in R ^{D} $, which pass through MICM and MLP separately to become $R ^{2} $ and we simultaneously supervise the outputs of MICM and MLP based on image-level annotation (tooth-marked or non-tooth-marked). 
In the inference stage, we multiply the MICM result by the mask on the edge of the tongue to obtain the recognition result.

\textbf{Multiple Instance Calculation Module (MICM).} The MICM is designed to effectively handle the output of the transformer encoder.
It treats each patch's feature representation as an instance within a bag, with the bag representing the entire tongue image.
If a bag contains at least one positive instance, the entire bag is considered positive.
Conversely, if all instances within a bag are negative, the bag is deemed negative.
Therefore, if a tongue image contains at least one tooth-marked instance, the image is classified as tooth-marked.
Therefore, we only need to focus on the most positive instance. 
The MICM performs several key operations: First, a multilayer perceptron (MLP) processes each instance, transforming feature representations from a $f \in R ^{BN \times D}$ tensor to a $f \in R ^{BN \times 2}$ tensor. This is followed by a Softmax operation, which is described by:
\begin{equation}
\label{eq:softmax}
\sigma(\mathbf{x})_i = \frac{e^{x_i}}{\sum_{j=1}^{N} e^{x_j}}.
\end{equation}

It converts the transformed features into a probability distribution, highlighting the regions most likely to contain tooth marks. 
Finally, the argmax function selects the instance with the highest probability, identifying the most relevant patch indicating tooth marks.
\begin{equation}
\label{eq:argmax}
\arg\max_xf(x)=\{x\mid\forall y:f(y)\leq f(x)\},
\end{equation}
where \( \mathbf{x} = [x_1, x_2, \ldots, x_n] \).
This approach leverages multiple instance learning to ensure the model focuses on the most critical parts of the image for accurate detection.
The selected most positive instance is represented as follows:
\begin{equation}
\label{eq:micm}
\hat{i} = \arg\max_{x_i \in \{x_1, \ldots, x_n\}} (\text{Softmax}(\text{MLP}(x_i))),
\end{equation}
where \( \{x_1, x_2, \ldots, x_n\} \) represents the instances, and $x_{\hat{i}}$ is the most positive instance selected. 
We then use this instance and the image label \( y \) to compute the supervised loss.

\textbf{Weakly Supervised Loss(WSL).} 
To supervise the learning process, we devised a weakly supervised loss function.
Initially, for \( y_{cls} \)  obtained directly from the MLP, we use the cross-entropy loss \( \mathcal{L}_{cls} \) (Eq.~\ref{eq:lcls}) for supervision.
\begin{equation}
\label{eq:lcls}
\mathcal{L}_{cls} = -\frac{1}{M}\sum_{i=1}^{M} y_i \log({y_{cls}}_i),
\end{equation}
here \( M \) is the number of samples, \( y_i \) represents the ground truth label, \( {y_{cls}}_i \) denotes the predicted probability
We simultaneously use cross-entropy loss and focal loss~\cite{lin2017focal} for $y_{\hat{i}}$ derived from MICM. 
Given the imbalanced distribution of tongue images with and without tooth marks in the datasets, we introduce focal loss to address the class imbalance. 
Focal loss reduces the weight of easy negative samples and focuses on fewer positive samples, which is crucial in medical image analysis where tooth-marked images are sparse and challenging to detect.
The formulation for the MIL loss is given by (Eq.~\ref{eq:lmil})
\begin{equation}
\begin{aligned}
\label{eq:lmil}
\mathcal{L}_{\text{mil}} = & -\frac{1}{M}\sum_{i=1}^{M} ( y_k \alpha_k (1 - y_{\hat{i}})^\gamma \log(y_{\hat{i}})
\\ & + \beta ( y_i \log (y_{\hat{i}}))),
\end{aligned}
\end{equation}
\( y_{\hat{i}} \) represents the predicted label probability. 
\( \alpha \) is a weighting factor to balance the importance of positive and negative examples. 
\( \gamma \) is the focusing parameter that adjusts the rate at which easy examples are down-weighted. 
\(\beta\) is a weight factor used to balance the cross-entropy loss and the focal loss.
In our experiments, these values are set to 0.25, 2.0, and 0.5 respectively. 
We have combined the focal loss component with the cross-entropy loss component within the \( \mathcal{L}_{mil} \) formulation. 
This allows simultaneous use of both loss functions for better handling of class imbalance and challenging samples in multiple instance learning frameworks.
The overall loss \( \mathcal{L}_{all} \) (Eq.~\ref{eq:lall}) is computed as the sum of \( \mathcal{L}_{cls} \) and a weighted combination of \( \mathcal{L}_{mil} \):
\begin{equation}
\label{eq:lall}
\mathcal{L}_{all} = \mathcal{L}_{mil} + \lambda \mathcal{L}_{cls}, 
\end{equation}
where \( \lambda \) is a weighting coefficient that balances the  \( \mathcal{L}_{cls} \) and\(\mathcal{L}_{mil} \) and is set to 0.5 in this paper.

\begin{table*}[!t]
\centering
\caption{Five-fold cross-validation experimental results on tooth-marked tongue datasets.}
\resizebox{0.8\linewidth}{!}{
\begin{tabular}{{l|cccc|cccc}}
\toprule
\multirow{2}{*}{Model} &  \multicolumn{4}{c|}{Our dataset} & \multicolumn{4}{c}{Public dataset} \\
& Accuracy & Precision & Recall & F1 Score & Accuracy& Precision & Recall & F1 Score \\
\midrule
Fold 1  & 84.0\% & 70.6\% & 82.4\% & 0.76 & 89.2\% & 84.0\% & 89.9\% & 0.87 \\
Fold 2  & 84.8\% & 86.5\% & 65.3\% & 0.74 & 85.6\% & 83.9\% & 83.9\% & 0.84 \\
Fold 3  & 81.7\% & 65.4\% & 81.7\% & 0.73 & 89.6\% & 86.7\% & 86.2\% & 0.88 \\
Fold 4  & 85.7\% & 80.5\% & 66.0\% & 0.73 & 91.2\% & 89.8\% & 89.8\% & 0.90 \\
Fold 5  & 85.1\% & 81.9\% & 75.4\% & 0.79 & 89.2\% & 88.9\% & 88.1\% & 0.89 \\
\midrule
Average & 84.2\% & 77.0\% & 74.2\% & 0.75 & 89.0\% & 86.7\% & 88.3\% & 0.87 \\
\bottomrule
\end{tabular}}
\label{tab:five_fold}
\end{table*}

\section{Experiment}
\subsection{Experiment setup}
\textbf{Datasets.} We used two datasets for our experiments. The primary dataset, collected by the Third Affiliated Hospital of Fujian University of Traditional Chinese Medicine, includes 1745 tongue images with an original resolution of 5184 $\times$ 3456 pixels, captured with professional medical equipment.
In the first stage of tongue detection, we annotated the bounding boxes of the tongue in 255 images. 
For the second stage, 556 images were labeled as tooth-marked tongues, and 1188 images were labeled as non-tooth-marked tongues. 
Two doctors from Fujian University of Traditional Chinese Medicine, each with over 10 years of experience, performed the annotations and cross-verified them. 
Additionally, we used publicly available datasets from Kaggle~\cite{tooth-marked-tongue} for training and evaluation.

\textbf{Implementation details.} We implemented our method using PyTorch and evaluated it on the aforementioned datasets. 
For the first stage of tongue detection,  we used 90\% of the annotated 255 images (90\% for training and 10\% for validating) to train a YOLOv8n model, and the remaining 10\% were used for testing. 
The images were resized to a resolution of 640 pixels, and the model was trained for 100 epochs. 
Detected and extracted tongue images were then resized to 224x224 pixels for the second stage. 
In this stage, we employed a ViT pre-trained on ImageNet for tooth-marked tongue recognition, removing its fully connected layer. 
The training was performed on a GeForce RTX 3090 GPU with a batch size of 32 and an initial learning rate of 0.00001. We used the AdamW optimizer and CosineAnnealingWarmRestarts learning rate scheduler, with 30 epochs.

\textbf{Evaluation Metric.} We used several metrics to evaluate our model, similar to other object detection methods~\cite{ren2015faster, redmon2016you, carion2020end} and weakly supervised tooth-mark tongue recognition approaches~\cite{zhou2022weakly, li2018tooth, sun2019tooth}. For tongue detection, we used Accuracy and Mean Average Precision. For tooth-marked tongue recognition, we used Accuracy, Precision, Recall, and F1 Score.
\begin{figure}[!t]
    \centering
    \includegraphics[width=0.49\linewidth]{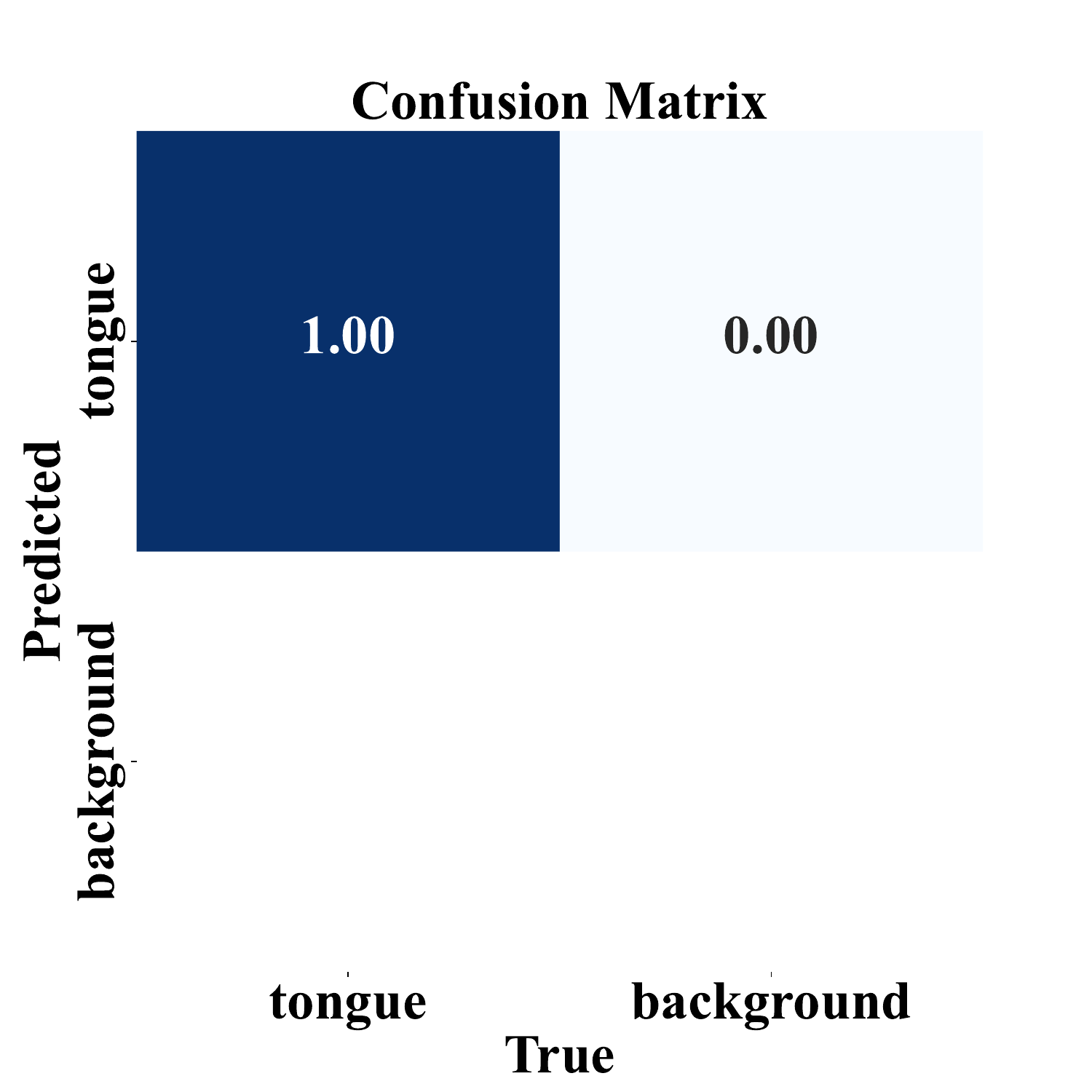}
    \includegraphics[width=0.49\linewidth]{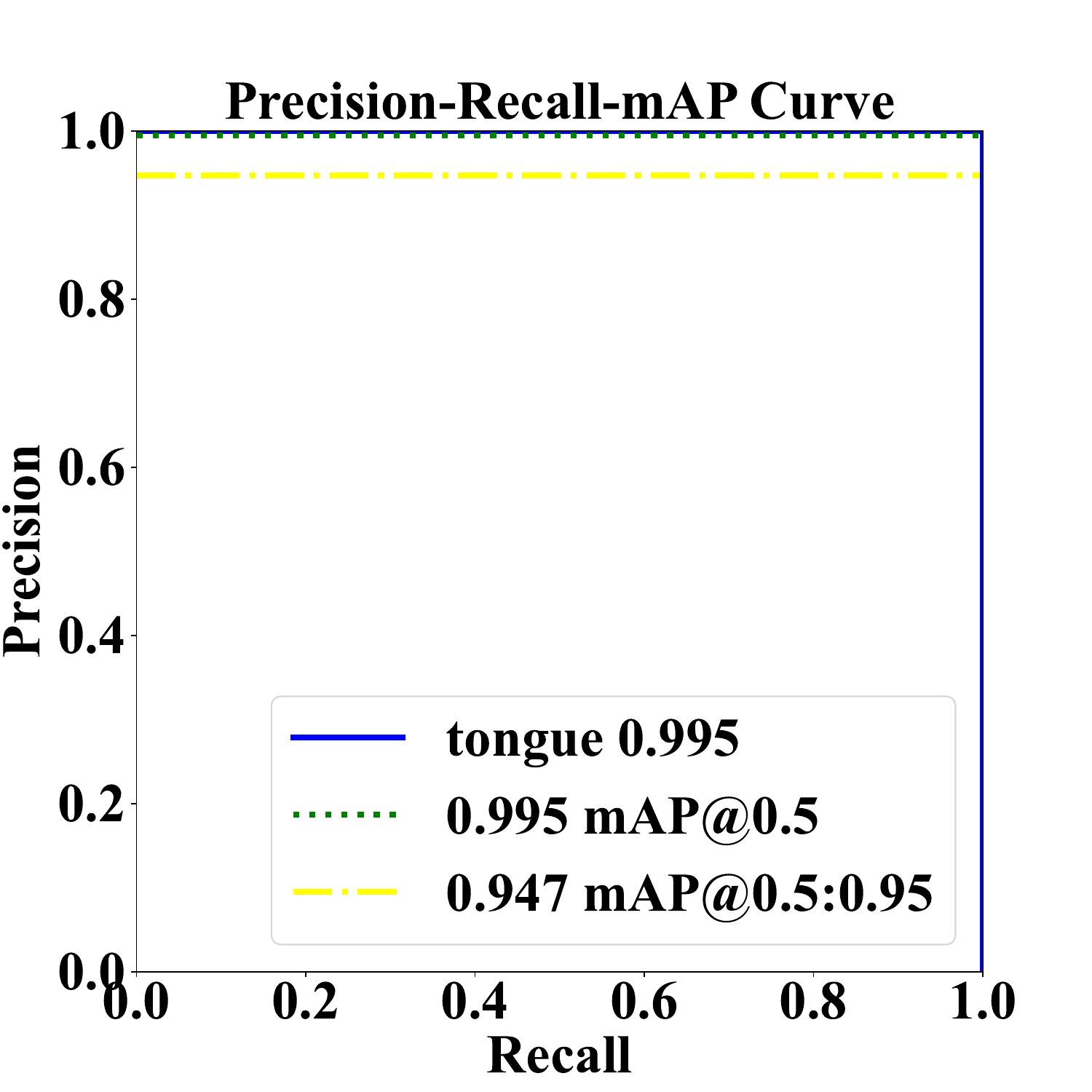}
    \caption{Perfomance of tongue detection using yolov8n}
    \label{fig:detect_result}
\end{figure}

\begin{figure}
    \centering
    \includegraphics[width=1\linewidth]{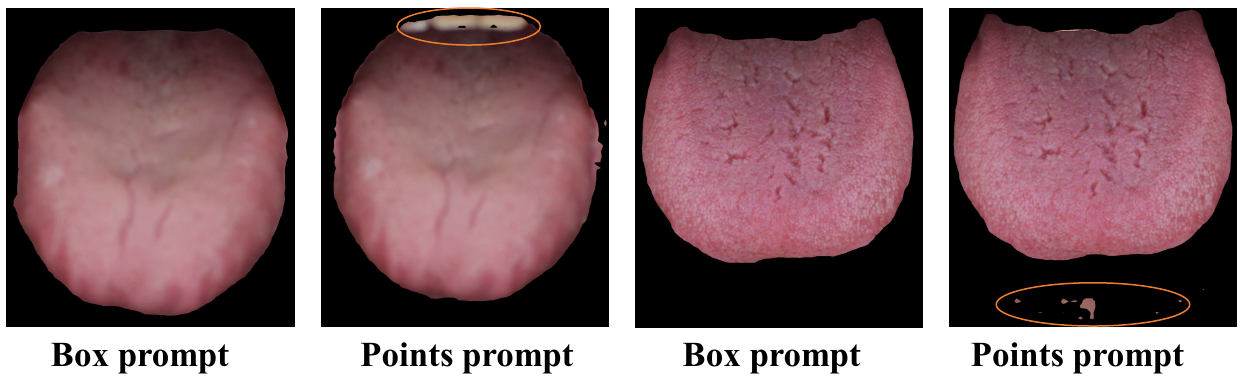}
    \caption{The impact of using boxes and points as prompts on scalloped tongue segmentation.}
    \label{fig:sam_tongue}
\end{figure}

\subsection{Tongue Detection Experiments}
We trained and tested our model on 255 manually annotated images. 
Our model quickly converged, and we selected the weights from the epoch with the best validation performance for testing.
The test results are shown in Fig.~\ref{fig:detect_result}. 
The YOLOv8n model for tongue detection achieved an accuracy and recall rate of 1.0, with mAP50 reaching 0.995 and mAP50-95 reaching 0.947.
These results demonstrate that our trained model performs strongly in tongue detection, providing highly accurate bounding boxes as prompts for the SAM.

We explored the performance differences between using points and boxes as prompts for the SAM.
We found that using boxes as prompts generally resulted in more accurate segmentation masks. 
As shown in Fig.~\ref{fig:sam_tongue}, we present two representative tongue segmentation results. 
When using points as prompts, the model sometimes segments teeth or other areas outside the tongue. 
In contrast, using boxes provides more comprehensive information, avoiding such issues.

\subsection{Weakly Supervised Recognition}
\subsubsection{Five Folds Cross Experiment}
We initially conducted experiments on two datasets to evaluate the performance of our proposed method. 
The results are presented in Table~\ref{tab:five_fold}. 
Our method achieves an average accuracy of 84.2\% on our private dataset and 89.0\% on the public dataset for tooth-marked tongue classification. 
Additionally, the precision, recall, and F1 scores remain stable, indicating the reliability of our results, with no extreme values in precision or recall.
Furthermore, in the five-fold cross-validation experiments, our method consistently maintained an accuracy above 80\% on the first dataset and above 85\% on the second dataset.
This demonstrates the robustness and stability of our model's performance across various experiments.
\begin{table}[!t]
\centering
\caption{Comparison Experiments on our dataset. We evaluated the performance of different methods under fair conditions. The asterisk `*' indicates that we modified the original conditions to ensure that the methods are consistent with others in terms of backbone and input size.}
\resizebox{1\linewidth}{!}{
\begin{tabular}{{l|cc|cccc}}
\toprule
Method & Backbone & Input size & Accuracy& Precision & Recall & F1 Score   \\ 
\midrule
\textcolor{gray}{\underline{\textit{W/o BBG}:}} & & & & & & 
\\
WTFF~\cite{tan2023tooth}   & ResNet34 & 224     & 72.8\% & 66.1\% & 42.2\%    & 0.47 \\ 
DCNN~\cite{tang2020automatic}   & ResNet34 & 416     & 80.7\% & 71.6\% & 65.3\%    & 0.68 \\ 
DCNN*  & ResNet34 & 224     & 81.4\% & 75.3\% & 61.9\%    & 0.68 \\ 
\midrule
\textcolor{gray}{\underline{\textit{W BBG}:}} & & & & & & 
\\
MILCNN~\cite{li2018tooth} & VGG16    & 227     & 60.1\% & 40.7\% & 54.9\%    & 0.43 \\
MILCNN*& ResNet34 & 224     & 57.6\% & 40.2\% & 51.4\%    & 0.37 \\ 
WSTDN~\cite{zhou2022weakly}  & ResNet34 & 224     & 79.2\% & 74.0\% & 56.7\%    & 0.59 \\ 
WSVM   & ViT-Base & 224     & \textbf{84.2\%} & \textbf{77.0}\% & \textbf{74.2}\%    & \textbf{0.75} \\ 
\bottomrule
\end{tabular}}
\label{tab:compare_ours}
\end{table}

\begin{table}[!t]
\centering
\caption{Comparison Experiments on Kaggle dataset. We conducted tooth-marked tongue classification experiments on the Kaggle dataset, using the same methods as presented in Table~\ref{tab:compare_ours}.}
\resizebox{1\linewidth}{!}{
\begin{tabular}{{l|cc|cccc}}
\toprule
Method & Backbone & Input size & Accuracy& Precision & Recall & F1 Score   \\ 
\midrule
\textcolor{gray}{\underline{\textit{W/o BBG}:}} & & & & & & 
\\
WTFF~\cite{tan2023tooth}   & ResNet34 & 224      & 80.9\% & 81.6\% & 75.2\% & 0.77 \\
DCNN~\cite{tang2020automatic}   & ResNet34 & 416      & 89.2\% & 86.7\% & 85.9\% & 0.86 \\ 
DCNN*  & ResNet34 & 224      & 86.0\% & 81.4\% & 83.8\% & 0.83 \\ 
\midrule
\textcolor{gray}{\underline{\textit{W BBG}:}} & & & & & & 
\\
MILCNN~\cite{li2018tooth} & VGG16    & 227      & 64.7\% & 56.4\% & 91.4\% & 0.70 \\
MILCNN*& ResNet34 & 224      & 57.0\% & 50.4\% & 92.4\% & 0.65 \\ 
WSTDN~\cite{zhou2022weakly}  & ResNet34 & 224      & 87.2\% & 81.5\% & \textbf{91.8}\% & 0.86 \\
WSVM   & ViT-Base & 224      & \textbf{89.0}\% & \textbf{86.7}\% & 88.3\% & \textbf{0.87} \\ 
\bottomrule
\end{tabular}}
\label{tab:compare_kaggle}
\end{table}

\subsubsection{Compared to Other Methods} 
We also conducted experiments comparing our approach with other existing tooth-marked tongue recognition methods to demonstrate the superiority of our proposed weakly supervised tongue recognition method based on ViT and MIL. 
The methods we compared include MILCNN~\cite{li2018tooth} , DCNN~\cite{tang2020automatic}, WSTDN~\cite{zhou2022weakly},  and WTFF~\cite{tan2023tooth}. 
Due to the lack of publicly available datasets used by MILCNN, DCNN, and WSTDN, we performed our experiments using our private dataset and a publicly available dataset from Kaggle.
Additionally, the code for the methods by MILCNN, DCNN, and WTFF is not released; therefore, we faithfully reimplemented these methods according to their respective publications and evaluated them on the datasets above. 
The comparative results of these experiments are presented in Table~\ref{tab:compare_ours} and Table~\ref{tab:compare_kaggle}. 
Among them, MILCNN* and DCNN* are the experiments we have been conducting to maintain backbone and image input size with other methods for fair comparison.

In Table~\ref{tab:compare_ours}, MILCNN, WSTDN, and our methods have the ability to generate bounding boxes for tooth-marked tongue regions (With bounding boxes generated(W BBG)). MILCNN method uses the concave regions of the tongue edge as proposals and extracts features using VGG16~\cite{simonyan2014very}. We also experimented with ResNet34 as the backbone.
These features are then classified using a multi-instance learning SVM. 
The WSTDN method uses uniformly sized rectangular regions along the tongue edge as proposals, with ResNet34~\cite{he2016deep} for both feature extraction and classification. 
DCNN and WTFF, however, do not use region proposals and rely solely on CNNs for classification. As a result, their recognition ability depends on additional methods like Grad-CAM, which cannot accurately identify bounding boxes(Without bounding boxes generated(W/o BBG)).
In contrast, our method utilizes ViT-Base as the backbone, with its patches serving as proposals. 
Our method achieved state-of-the-art results on our private dataset, with an accuracy of 2.8 percentage points higher than previous methods.
Additionally, our method significantly outperforms other recognition methods in precision, recall, and F1 score.
This performance improvement benefits from the Vision Transformer’s ability to capture comprehensive global information and facilitate information exchange between image patches. 
Table~\ref{tab:compare_kaggle} demonstrates that our method also performs excellently on the Kaggle black-and-white dataset. 
Our method's accuracy is 0.2 percentage points lower than that of purely classification-based DCNN. In the W BBG group, our recall rate is lower than that of WSTDN because our WSL focuses more on recognizing positive samples (tooth-marked tongues), resulting in our accuracy and precision exceeding those of WSTDN by 1.8\% and 5.2\%, respectively.
Overall, our method exhibits robust performance in distinguishing tooth-marked tongue images. 
This indicates our ability to effectively recognize and identify tooth-marked regions.

\subsubsection{Cracked Tongue Recognition}
To explore the generalizability of our approach, we conducted experiments on classifying cracked tongues to validate our method's applicability across various weakly supervised recognition tasks.
The region proposals used by MILCNN and WSTDN were specifically designed for tooth-marked tongue detection and are not applicable to cracked tongue recognition.
Additionally, DCNN and WTFF only possess classification capabilities without explicit recognition features. 
Consequently, we compared our method against standard ResNet34 and ViT-Base models. 
The results are shown in Table~\ref{tab:crack}. 
Using ViT-Base directly results in a 2.6 percentage point higher accuracy compared to ResNet34, indicating that ViT-Base has a stronger classification ability than ResNet.
Moreover, our method improves accuracy by an additional 2.7 percentage points over ViT-Base, demonstrating the significance of our MICM and WSL.
Our accuracy and precision are higher than those of ViT-Base, suggesting that the focal loss we employed is effective.

\begin{table}[!t]
\vspace{1.2em}
\centering
\caption{Comparison of Cracked Tongue Classification. We compared the performance of our method with ResNet34 and ViT-Base for cracked tongue classification.}
\resizebox{0.9\linewidth}{!}{
\begin{tabular}{{c|cccc}}
\toprule
Method  & Accuracy& Precision & Recall & F1 Score   \\ 
\midrule
ResNet34  & 86.8\% & 72.9\% & 86.0\%  & 0.79  \\ 
ViT-Base  & 89.4\% & 88.3\% & \textbf{93.7}\% & \textbf{0.92} \\ 
WSVM      & \textbf{91.7}\% & \textbf{89.4}\% & 85.6\% & 0.87 \\ 
\bottomrule
\end{tabular}}
\label{tab:crack}
\end{table}

\begin{table}[!t]
\centering
\caption{Ablation Study. We compared the performance of models with and without the Multiple Instance Calculation Module (MICM) and the Weakly Supervised Loss (WSL) under consistent experimental conditions to ensure the effectiveness of these additions.}
\resizebox{1\linewidth}{!}{
\begin{tabular}{{ccc|cc|c}}
\toprule
\multirow{2}{*}{CLS} & \multirow{2}{*}{MICM} & \multirow{2}{*}{WSL} & \multicolumn{2}{c|}{Tooth-marked Tongue}  & \multirow{2}{*}{Cracked Tongue}  \\ \cmidrule(lr){4-5}  
 &   &  &  Ours & Public & \\
\midrule
$\checkmark$   &   &  & 82.7\% & 88.5\% & 89.4\% \\
& $\checkmark$    &   & 81.8\% & 88.5\% &  90.9\%\\
$\checkmark$ & $\checkmark$  &  & 83.4\% &\textbf{ 89.1\%} & 90.0\% \\
$\checkmark$ & $\checkmark$ & $\checkmark$ & \textbf{84.2\%} & 89.0\%  & \textbf{91.7\%}   \\
\bottomrule
\end{tabular}}
\label{tab:ablation}
\end{table}

\begin{figure*}[t]
\begin{center}
\includegraphics[width=0.93\linewidth]{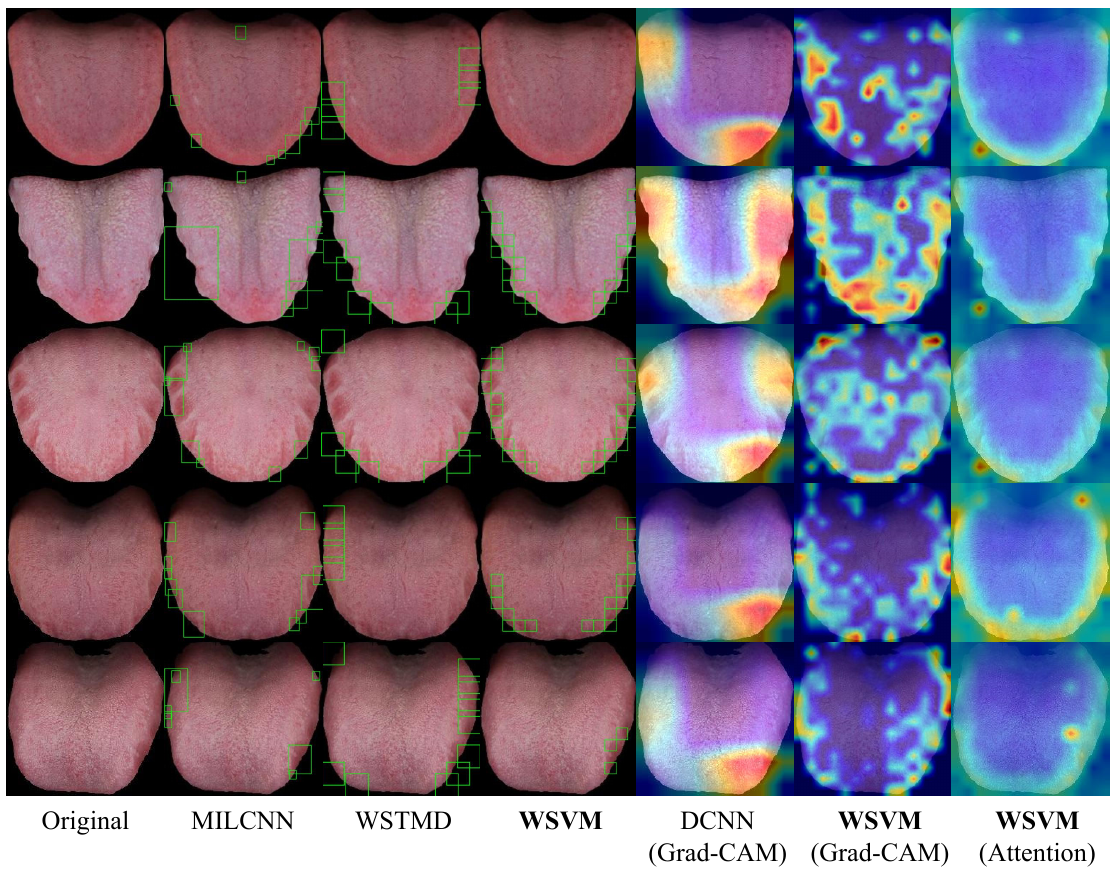}
\end{center}
\caption{Visualization Results. The first row shows results for non-tooth-marked tongues, while the second to fifth rows display recognition results for tooth-marked tongues. Among the various recognition methods, MILCNN, WSTDN, and our proposed method generate final bounding boxes based on region proposals. For comparison with DCNN, we visualized the Grad-CAM and boundary attention results of our method.}
\label{fig:visual}
\end{figure*}

\subsection{Ablation Study}
We conducted ablation experiments to investigate the effectiveness of our proposed multiple instance calculation module and the designed weakly supervised loss. 
As shown in Table~\ref{tab:ablation}, using a standard Vision Transformer (ViT) for direct tooth-marked tongue classification yields an accuracy of only 82.7\% and lacks clear recognition capabilities. 
Using only MICM results in a decrease in accuracy for WSVM. However, combining both the CLS and MICM models provides approximately a 0.6-point improvement across three tasks. 
Adding WSL further enhances our method, achieving improvements of 0.8 and 1.7 points on our Tooth-marked Tongue and Cracked Tongue datasets, respectively, while the change is minimal on the publicly available Tooth-marked Tongue dataset.
This may be because the publicly available dataset (546 marked and 704 unmarked) is more balanced compared to our datasets (556 marked and 1188 unmarked; 602 cracked and 1142 untracked).
In summary, our MICM and WSL methods enhance the model's ability to predict image patches corresponding to tooth-marked and cracked regions, improving the recognition capabilities of the weakly supervised model.

\subsection{Visualization}

We conducted visualization experiments to demonstrate WSVM's recognition capability. 
As shown in Fig.~\ref{fig:visual}, in the first row, in the first row, which displays the results for non-tooth-marked tongues, WSVM is the only one that correctly avoids generating bounding boxes for tooth-marked regions, unlike MILCNN and WSTDN. 
In the second row, we selected a tooth-marked tongue with prominent indentations along the edges. Both MILCNN and WSVM, which generate region proposals based on these indentations, successfully captured the tooth marks on both sides of the tongue. 
However, WSTDN failed to detect the marks on the right side of the tongue. The third row features a tooth-marked tongue without noticeable indentations but with distinct color characteristics along the edges. 
Here, WSVM outperformed MILCNN and WSTDN by more comprehensively identifying the tooth-marked areas on both sides of the tongue. The fourth and fifth rows present tooth-marked tongues with small and less discernible tooth marks. 
WSVM accurately located these small tooth-marked regions. In contrast, MILCNN produced numerous overly small and overlapping boxes, while WSTDN identified regions that were far from the actual tooth marks. WSVM effectively addressed these issues. 
Additionally, to compare with DCNN, which does not generate bounding boxes, we used Grad-CAM to visualize the indicative regions identified by our model during tooth-marked tongue classification. 
We also employed Attention Rollout to visualize the edge attention of our model. 
Compared to DCNN, our model demonstrated the ability to recognize more subtle areas, with attention concentrated on the tongue edges, which are more indicative of tooth marks.

\section{Conclusion}
In this work, we proposed an automatic two-stage method for recognizing tooth-marked tongues named WSVM, combining fully automatic tongue extraction and a weakly supervised recognition framework. 
In the first stage, we used YOLOv8n for tongue detection and the Segment Anything Model (SAM) for precise segmentation, achieving near-perfect accuracy and recall.
The second stage employs a ViT integrated with a Multiple Instance Calculation Module (MICM) to recognize tooth-marked regions effectively. 
Extensive experiments show the superiority of WSVM over the competing state-of-the-art. 
In addition, WSVM can be effectively applied to other weakly supervised recognition tasks in tongue diagnosis, such as cracked tongue recognition.
Our work offers a reliable tool for automated tooth-marked tongue diagnosis, enhancing clinical assessment and treatment capabilities.
Future research can extend WSVM to other tongue diagnosis applications, furthering its clinical utility.

\section*{Acknowledgement}
This work is supported by National Key R\&D Program of China (No. 2023YFC3503004); Traditional Chinese Medicine Foundation of Xiamen (No. XWZY-2023-0623); Scientific Research Foundation for the High-level Talents, Fujian University of Traditional Chinese Medicine (No.X2020008-talents); Science and Technology Planning Project of Fujian Province of China (No. 2021I0018); and 2023 Basic Disciplines Research Enhancement Program, Fujian University of Traditional Chinese Medicine (No. XJC2023003).

\bibliographystyle{cas-model2-names}

\bibliography{cas-refs}

\end{document}